\documentclass{article}

\usepackage{tikz}
\usepackage{microtype}
\usepackage{graphicx}
\usepackage{subfigure}
\usepackage{booktabs} 

\usepackage{hyperref}
\usepackage{url}

\usepackage[accepted]{icml2025}
\usepackage{adjustbox}

\usepackage{amsmath}
\usepackage{amssymb}
\usepackage{mathtools}
\usepackage{amsthm}
\usepackage{arydshln}

\usepackage{multirow} 

\usepackage{CJKutf8} 

\usepackage[capitalize,noabbrev]{cleveref}

\theoremstyle{plain}

\theoremstyle{definition}

\theoremstyle{remark}

\usepackage[textsize=tiny]{todonotes}

\graphicspath{{figures/}{}}

\icmltitlerunning{Retrieval Backward Attention without Additional Training: Enhance Embeddings of Large Language Models via Repetition.}

\begin{document}

\twocolumn[
\icmltitle{Retrieval Backward Attention without Additional Training: Enhance Embeddings of Large Language Models via Repetition.}

\begin{icmlauthorlist}
\icmlauthor{Yifei Duan}{bnu}
\icmlauthor{Raphael Shang}{waiyan}
\icmlauthor{Deng Liang}{waiyan}
\icmlauthor{Yongqiang Cai}{bnu}
\end{icmlauthorlist}

\icmlaffiliation{bnu}{School of Mathematical Sciences, Laboratory of Mathematics
and Complex Systems, MOE, Beijing Normal University, Beijing,
100875, China.}

\icmlaffiliation{waiyan}{Beijing Waiyan Online Digital Technology Co., Ltd, Beijing, China}

\icmlcorrespondingauthor{Yongqiang Cai}{caiyq.math@foxmail.com}
\icmlkeywords{Machine Learning, ICML}

\vskip 0.3in
]

\printAffiliationsAndNotice{} 

\begin{abstract}
Language models can be viewed as functions that embed text into Euclidean space, where the quality of the embedding vectors directly determines model performance, training such neural networks involves various uncertainties. This paper focuses on improving the performance of pre-trained language models in zero-shot settings through a simple and easily implementable method. We propose a novel backward attention mechanism to enhance contextual information encoding. Evaluated on the Chinese Massive Text Embedding Benchmark (C-MTEB), our approach achieves significant improvements across multiple tasks, providing valuable insights for advancing zero-shot learning capabilities.
\footnote{Our code is available at \url{https://github.com/cqdyf099/ReBA}}
\end{abstract}
\icmlkeywords{Attention Mechanism, Context Information, Echo Encoding, Bidirectional Attention}

\section{Introduction}


Text embedding learning is a key task in Natural Language Processing (NLP) that transforms text into vector representations, making them computationally tractable. Formally, given a text $x$ in a corpus space $X$, embedding learning aims to train a model $ f: X \times \theta \to \mathbb{R}^d $ to generate continuous vector representations $ v = f_{\theta}(x) $ that capture semantic information and enhance performance in downstream tasks such as information retrieval (IR), semantic similarity estimation, classification, and clustering \cite{ni2021large,muennighoff2022mteb}.

\begin{figure}
    \centering
    \includegraphics[width=0.8\linewidth]{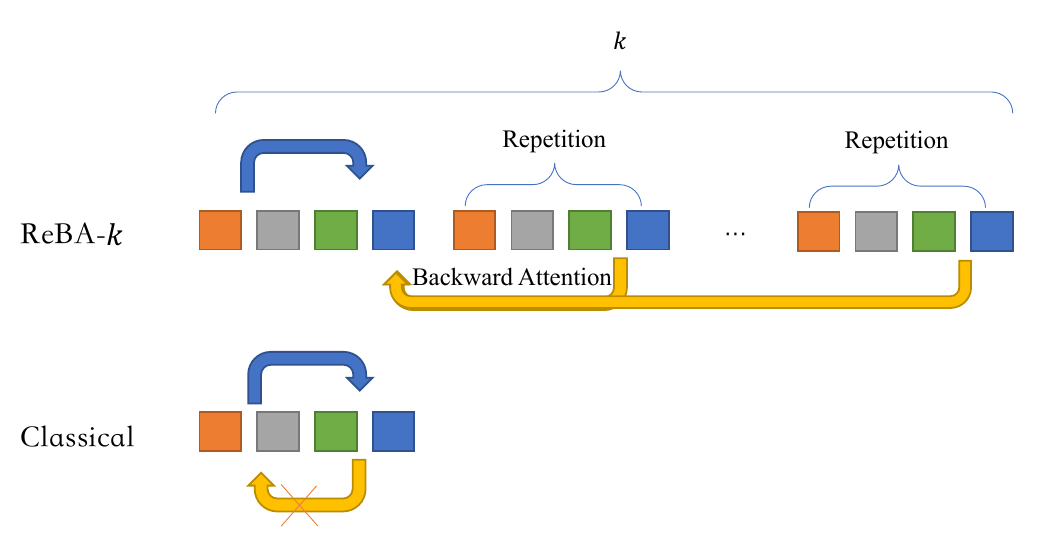}
    \caption{Illustration of ReBA embedding: The classical embedding method captures only the contextual information preceding the token. In contrast, ReBA enhances the quality of target token embeddings by repeating the text \(k-1\) times, computing a weighted sum of the target token's original embedding and subsequent token embeddings using backward attention weights. When the sentence appears \(k\) times, the resulting embeddings are referred to as ReBA-\(k\) embedding.}
    \label{fig:ReBA_intro}
\end{figure}

Advanced large language models (LLMs) have recently demonstrated exceptional generalization and transfer capabilities across downstream tasks. However, Transformer-based models like GPT \cite{radford2018improving} (unidirectional) and BERT \cite{kenton2019bert} (bidirectional) are often specialized in different tasks due to the influence of pretraining tasks, with BERT excelling in Natural Language Understanding (NLU) and GPT excelling in Natural Language Generation (NLG), this results in the need to maintain multiple independent models \cite{dong2019unified}. Research shows that decoder-only models, pre-trained with next-token prediction, achieve superior zero-shot generalization performance across various tasks \cite{ wang2022language}, while its performance on NLU is still not good as that of bidirectional models, some prompt methods can be implemented to improve the embedding quality \cite{jiang2023scaling,liu2024gpt}, in this paper, we also focus on improving the embedding quality of decoder-only models from the perspective of the input text by \emph{repetition} and \emph{backward attention} and we call it ReBA embedding (Figure \ref{fig:ReBA_intro}).


\textbf{Repetition:} We focus on improving the performance of decoder-only models in a zero-shot setting. Research shows that repeating input sentences can provide additional contextual information, significantly boosting model performance. For instance, \cite{jelassirepeat} found that text repetition allows Transformer models to outperform state-space models like S4 \cite{gu2021efficiently} and Mamba \cite{gu2023mamba}. Similarly, \cite{arora2024just,springer2025repetition} demonstrated quality improvements in language models through repeated contexts. While repetition enhances decoder-only models, their performance still lags behind bidirectional models (see Table \ref{tab:main_results} and Figure \ref{fig:word_results}). To address this gap, we propose a simple backward attention mechanism to further improve context encoding. Although our method remains inferior to bidirectional models, it achieves substantial improvements over simple repetition and classical embeddings in zero-shot scenarios, offering valuable insights for future research.

\textbf{Backward Attention:} 

In decoder-only architectures, forward attention is typically used, where the attention matrix is a lower triangular matrix. This design lacks associations with subsequent context. Backward attention, on the other hand, represents the relationship between a token and its subsequent context, as reflected in the attention matrix. To enhance the encoding quality of the original text, we propose leveraging repeated text. By concatenating \(x\) with itself, represented as \(x + x'\) using string addition, the embeddings of \(x'\) are used to enhance the embeddings of \(x\). Since \(x'\) appears after \(x\), it is natural to consider using backward attention to strengthen \(x\)'s encoding quality.

In existing models, there are bidirectional language models such as BERT \cite{kenton2019bert}, RoBERTa \cite{liu2019roberta}, XLNet \cite{yang2019xlnet}, and other models like \cite{jiao2019tinybert,clark2020electra}. Their attention matrices are symmetric (Figure \ref{fig:attention_relationship}). Encoder-decoder models, such as T5 \cite{raffel2020exploring}, do not have strictly symmetric attention structures, but they enable the model to attend to both preceding and subsequent context. Although these models employ bidirectional or partially bidirectional structures, introducing new text alters the embeddings of all preceding tokens, significantly increasing computational costs. 

\begin{figure}[h!]
    \centering

    \begin{tikzpicture}[scale=1, every node/.style={font=\small}]
        \foreach \i in {1,2,3,4} {
            \node[circle, draw, minimum size=0.5cm] (t\i) at (\i*1.5, 0) {$w_\i$};
            \node[circle, draw, minimum size=0.5cm, fill=blue!20] (e\i) at (\i*1.5, 1.5) {$v_\i$};
        }

        \foreach \i in {1,2,3,4} {
            \foreach \j in {1,2,3,4} {
                \draw[->, blue!50] (t\j) -- (e\i);
            }
        }
        \node at (3, 2.8) {BERT (Bidirectional Attention)};

        \begin{scope}[shift={(7.5, -0.5)}]
            \node at (1, 2.5) {symmetric};
            \draw[step=0.5, black!100] (0,0) grid (2,2);
            \foreach \x in {0,...,3} {
                \foreach \y in {0,...,3} {
                    \ifnum \x=\y
                        \fill[blue!50] (\x/2,1.5-\y/2) rectangle (\x/2+0.5,1.5-\y/2+0.5);
                    \else
                        \fill[blue!30] (\x/2,1.5-\y/2) rectangle (\x/2+0.5,1.5-\y/2+0.5);
                    \fi
                }
            }
        \end{scope}
    \end{tikzpicture}

    \vspace{1cm}

    \begin{tikzpicture}[scale=1, every node/.style={font=\small}]
        \foreach \i in {1,2,3,4} {
            \node[circle, draw, minimum size=0.5cm] (t\i) at (\i*1.5, 0) {$w_\i$};
            \node[circle, draw, minimum size=0.5cm, fill=red!20] (e\i) at (\i*1.5, 1.5) {$v_\i$};
        }

        \foreach \i in {1,2,3,4} {
            \foreach \j in {1,...,\i} {
                \draw[->, red!50] (t\j) -- (e\i);
            }
        }
        \node at (3, 2.8) {GPT (Unidirectional Attention)};

        \begin{scope}[shift={(7.5, -0.5)}]
            \node at (1, 2.5) {lower triangle};
            \draw[step=0.5, black!100] (0,0) grid (2,2);
            \foreach \x in {0,...,3} {
                \foreach \y in {0,...,\x} {
                    \fill[red!30] (\y/2,1.5-\x/2) rectangle (\y/2+0.5,1.5-\x/2+0.5);
                }
            }
        \end{scope}
    \end{tikzpicture}
    \caption{Illustration of Attention Relationships in BERT (Bidirectional Attention) and GPT (‌Causal Attention) with Corresponding Attention Matrix Representations}
    \label{fig:attention_relationship}
\end{figure}

In contrast, decoder-only models such as the GPT series \cite{radford2018improving,radford2019language,brown2020language}, the LLaMA series \cite{touvron2023llama,touvron2023llama2}, Qwen \cite{bai2023qwen}, and Baichuan \cite{baichuan2023a} use lower triangular attention matrices (Figure \ref{fig:attention_relationship}). This means that each token can only attend to preceding tokens. Adding new tokens after a token does not change its embedding, reducing computational costs. This characteristic provides an excellent opportunity for us to selectively enhance the embeddings of specific tokens using backward attention. And our contributions are as follows:
\begin{itemize}
    \item We propose a novel algorithm that significantly enhances the embedding quality of pretrained models, leading to improved performance in downstream tasks and stronger natural language understanding capabilities. Our method achieves significant improvements on multiple tasks.
	\item The method achieves these improvements without requiring additional training of the model, new models, or extra parameters, ensuring simplicity and efficiency.
	\item 
	Our algorithm maintains the advantages of unidirectional LLMs while capturing subsequent context information, enabling targeted enhancement of the embeddings of specific tokens.


\end{itemize}

\section{Preliminaries}

Our goal is to obtain an embedding vector $LLM(C)\in \mathbb{R}^d$ for a sentence $C$ and embedding vector $LLM(w|C)\in \mathbb{R}^d$ for word $w$ in context $C$. This vector can be viewed as the semantic representation of the sentence or word, and our aim is to obtain a better semantic representation that can more effectively capture the semantic information of the sentence or word. These vectors can measure the similarity between sentences or words, and can be used to complete downstream tasks such as text classification, word sense disambiguation. 

We are particularly interested in extracting these vectors using an \emph{autoregressive language model}. An autoregressive language model predicts the next token in a sequence based on the preceding tokens. This mechanism inherently limits the model to capturing information only from earlier tokens, as reflected in the attention matrix—a lower triangular matrix where each token attends only to its preceding tokens. To enhance the embedding quality, we aim to reutilize the information in the attention matrix to also capture insights from subsequent tokens.

\subsection{Looking Back: Bidirectional Applications of Attention Mechanism}
\label{sec:bidirectional_attention}

The attention mechanism \cite{vaswani2017attention} can be understood mathematically as a mapping from a query vector $\mathbf{q}_j \in \mathbb{R}^d$ to a weighted sum of a set of key-value pairs $(\mathbf{k}_i, \mathbf{v}_i)$, where $ \mathbf{k}_i, \mathbf{v}_i \in \mathbb{R}^d$, $i$ indexes over the tokens in the input sequence. Formally, given an input sequence of embeddings $\{\mathbf{x}_1, \cdots, \mathbf{x}_n\}$, the attention mechanism computes the output embedding $v_j$ for each token as follows:

\begin{equation}
    v_j = \sum_{i=1}^j \alpha_{i,j} \mathbf{v}_i,
\end{equation}

where the attention weights $\alpha_i\in \mathbb{R}$ are derived from the query and key vectors using a compatibility function, typically the scaled dot-product:

\begin{equation}
    \alpha_{i,j} = \frac{\exp(\mathbf{q}_j \cdot \mathbf{k}_i / \sqrt{d})}{\sum_{m=1}^n \exp(\mathbf{q}_j \cdot \mathbf{k}_m / \sqrt{d})}.
\end{equation}

Here, $\mathbf{q}_j = \mathbf{W}_q \mathbf{x}_j, \mathbf{k}_i = \mathbf{W}_k \mathbf{x}_i,$ and $\mathbf{v}_i = \mathbf{W}_v \mathbf{x}_i$ are the query, key, and value vectors, obtained by trainable linear transformations $\mathbf{W}_q,\mathbf{W}_k,\mathbf{W}_v \in \mathbb{R}^{d\times d}$ of the input embeddings, and $d$ is the dimensionality of the query/key space.

In Transformer-based models, each token in the sequence simultaneously acts as a query, key, and value, resulting in contextualized embeddings for all tokens. This mechanism enables the model to integrate information across the entire sequence. For example, given an input sequence $\{w_1, \cdots, w_n\}$, the output embedding $v_n$ for the last token $w_n$ is computed as:

\begin{equation}
    v_n = \text{Attention}(\mathbf{q}_n, \{\mathbf{k}_1, \cdots, \mathbf{k}_n\}, \{\mathbf{v}_1, \cdots, \mathbf{v}_n\}),
\end{equation}

where $\mathbf{q}_n$ is derived from $w_n$ and $\{\mathbf{k}_i, \mathbf{v}_i\}$ are derived from all preceding tokens $\{w_1, \cdots, w_n\}$.

This computation illustrates how attention integrates global context. Notably, in autoregressive models, the attention mechanism is constrained such that $v_i$ only depend on $w_1, \cdots, w_i,$ ensuring that information flows unidirectionally. Such a framework blurs the traditional boundary between word and sentence embeddings. The embedding $v_n$ for the final token incorporates contextual information from the entire sequence, making it a natural representation for the full sentence. This marks a fundamental departure from traditional word embedding models like Word2Vec\cite{mikolov2013efficient}, which lack explicit mechanisms for modeling token interactions, as discussed in the next section.








\subsection{Word embedding and sentence embedding, are they the same?}


In traditional approaches like Word2Vec, the methods for obtaining sentence embeddings and word embeddings were notably different. This is because words had weak interdependencies, with each word’s embedding being relatively independent. Consequently, additional processing, such as average pooling or new sentence embedding methods (e.g., Sent2Vec \cite{moghadasi2020sent2vec}), was required to derive a meaningful sentence vector because one word’s embedding could not effectively represent the entire sentence.

However, large language models (LLMs) based on Transformer architectures and attention mechanisms differ fundamentally from traditional word embedding models. Each token’s embedding in these models is derived through interactions with other tokens. For example, in an autoregressive model, given an input sequence of tokens $\{w_1, \cdots, w_n\}$, the output embeddings $\{v_1, \cdots, v_n\}$ are generated. While $v_n$ can be viewed as the embedding for $w_n$, it can also be considered the embedding for the entire sentence. This is because $v_n$, influenced by the attention mechanism, incorporates information from the entire sentence. However, due to the unidirectional nature of autoregressive models, embeddings $v_1, \cdots, v_{n-1}$ do not include information from $v_n$.

Thus, while traditional static word embedding models, such as Word2Vec\cite{mikolov2013efficient} and Glove\cite{pennington2014glove}, demonstrate significant differences between word and sentence embeddings due to the lack of interaction between words in a context, Transformer-based models blur this distinction. With attention mechanisms, the last-token embedding can also effectively represent the entire sentence.


In this paper, we evaluate the performance of our method on word and sentence embeddings. To assess the quality of word embeddings, we examine the algorithm’s performance on word sense disambiguation datasets. To evaluate the quality of sentence embeddings, we consider the algorithm’s performance on the C-MTEB.

\subsection{Language model embedding}

We first extract embeddings from the activations of the final hidden layer of the language model. 
Given a sentence $C = \{w_1, \cdots, w_n\}$, we extract the embeddings of the tokens $w_i$ in $C$ from the model as $LLM(w_i| C)$ as the word embedding. 

In practice, we consider two main methods for sentence embedding \cite{wang2023improving,reimers2019sentence}, One approach called \textbf{last-token pooling} is to use the embedding of the last token as the sentence embedding: $LLM(w_{-1}|C)$, while the other called \textbf{mean token pooling} involves averaging the embeddings of all tokens to obtain the sentence embedding: $\frac{1}{|C|}\sum_{w\in C} LLM(w|C)$. In this paper, we adopt these methods to obtain sentence embeddings.


\subsection{New embedding via repetition and backward attention}

To enhance the natural language understanding capabilities of autoregressive models while avoiding the comprehensive update of all token embeddings as seen in bidirectional models when new tokens are added, we propose a novel method called \emph{ReBA} (Retrieval Backward Attention) embedding. This approach leverages the model’s inherent capabilities to more effectively capture bidirectional information and enhance model performance. 

\section{Main Method}

The core concept of ReBA involves repeating the input text \emph{twice} and extracting the attention matrix from the model. This attention matrix is then used to apply backward attention, updating the embeddings of specific tokens based on the repeated text. Finally, the updated embeddings are combined with the original embeddings to produce the final embedding vectors.

\subsection{Classical embedding ignore bidirectional context}

As discussed in Sec \ref{sec:bidirectional_attention}, classical sentence embeddings fail to effectively capture bidirectional information. In autoregressive language models, the contextualized embedding at position $k$ encodes information only from tokens preceding $k$, without considering tokens that follow. As a result, the tokens at the beginning of a sentence may fail to fully capture their intended meaning due to the lack of semantic information from the subsequent context.

\subsection{Repetition capture bidirectional context}

Research shows that text repetition can significantly enhance the bidirectional information captured by sentence embeddings in autoregressive models\cite{springer2025repetition,jelassirepeat}. By repeating text, the model’s embeddings become more contextually enriched. Instead of using a prompt like "Rewrite the sentence: $x$, rewritten sentence: $x$" in \cite{springer2025repetition}, we repeat the sentence directly to eliminate prompt effects. Taking "I love NLP." as an example, the repeated sentences are:
\begin{itemize}
    \item Repeated once: 'I love NLP. I love NLP.'
    \item Repeated twice: 'I love NLP. I love NLP. I love NLP.'
\end{itemize}
For the word 'love', we observe that the second and third occurrences of 'love' carry more contextual information than the first, as they capture the broader context. We adopt the term from \cite{springer2025repetition} and refer to these enhanced embeddings as \emph{Echo embeddings}.

\subsection{ReBA Embedding}

Inspired by the benefits of text repetition, we propose a new method that does not require the use of prompts in \cite{springer2025repetition}. Instead, we directly repeat the text and introduce backward attention to rely more on the model itself, reducing the introduction of additional parameters. Our method has three steps:

\subsection{First step: Construct Attention Matrix Extraction}
\subsubsection{Motivation}


\cite{vig2019analyzing} found that the performance of GPT2's multi-head attention matrices varies across different attention heads and hidden layers, and some of the attention heads in the deeper layers contribute less to the model's performance\cite{he2024matters}. Inspired by this observation, we aim to integrate this information and explore a new method to leverage all attention matrices to enhance the quality of embedding vectors.

To enhance the model's understanding of contextual relationships, we first extract the \textbf{attention matrix}. Specifically, we focus on the maximum attention value across all layers and heads of the transformer model. For each token pair, if any attention head in any layer assigns significant attention, we record this interaction in the attention matrix. This matrix effectively captures the most prominent token relationships identified by the model, irrespective of the layer or head.


Taking LLaMA-2-7B as an example, which has 32 attention heads and 32 hidden layers, we denote all attention matrices as $A^{p,q}\in \mathbb{R}^{n\times n}$, where $p$ represents the $p$-th attention head, $q$ represents the $q$-th hidden layer, with $n$ being the length of the input sequence. 


\subsubsection{methods}

This method integrates attention weights across multiple layers to construct a comprehensive symmetric attention matrix $A$ . The process involves the following steps:

	1.	Layer-wise Attention Integration:
For each layer $q$ of the transformer model, the output provides a set of attention heads represented as matrices $A^{\cdot,q}$. These matrices capture directional attention weights for tokens in the input sequence and are lower triangular matrix with all elements above the diagonal being zero.

	2.	Symmetrization:
To preserve the bidirectional attention between tokens, each attention matrix $A^{p,q}$ is converted into a symmetric form:

$$\tilde{A}^{p,q} = \frac{A^{p,q} + (A^{p,q})^T}{2}$$

3.	Iterative Fusion with Maximum Update Rule:
Initialize $A^{new} = 0$, the symmetric attention matrices from all layers $p$ and heads $q$ are fused iteratively using a “maximum update rule”:

$$A^{new} = \frac{A^{new} + \tilde{A}^{p,q}}{2} + \frac{|A^{new} - \tilde{A}^{p,q}|}{2},$$

this update ensures that the resulting matrix $A^{new}$ contains the maximum value of attention matrices among all layers and all heads. This matrix represents a global view of token interactions, integrating information across layers and attention heads in a symmetric and robust manner.

This approach effectively aggregates layer-wise attention dynamics into a unified, interpretable representation of token relationships, the details are presented in Algorithm \ref{alg:attention-matrix-extraction}.

\begin{algorithm}
    \caption{Attention Matrix Extraction}
    \textbf{Input:} Text with length $n$, pretrained language model $LLM$ and its number of hidden layers $I$, number of attention heads $J$.\\
    \textbf{Output:} New Attention Matrix $A^{new}$.\\
    1. Extract all attention matrices $A^{p,q} \in \mathbb{R}^{n \times n} \{p=1,\cdots,I; q=1,\cdots,J\}$ from the pretrained model, initialize $A^{new}=0$.\\
    2. \textbf{for} $p$ = $1,2,\cdots,I$: \par
        \hspace{2em} \textbf{for} $q$ = $1,2,\cdots,J$: \par
        \hspace{4em} $\tilde{A}^{p,q} = \frac{A^{p,q} + (A^{p,q})^T}{2}$ \# Symmetry \par
        \hspace{4em} $A^{new} = \frac{A^{new} + \tilde{A}^{p,q}}{2} + \frac{|A^{new} - \tilde{A}^{p,q}|}{2}$ \# Max attention.\\
    3. \textbf{Return} $A^{new}$.
    \label{alg:attention-matrix-extraction}
\end{algorithm}

\subsection{Second step: Backward Attention and Text Repetition}

To compute text embeddings effectively, we integrate \emph{text repetition} and a \emph{backward attention} mechanism as outlined in Algorithm \ref{alg:word_embedding_construction} (for word embedding) and Algorithm \ref{alg:embedding-construction} (for sentence embedding).

Given a text sequence $ C = \{w_1, w_2, \dots, w_n\} $, we first apply \emph{text repetition} by duplicating the sequence to form a new input:  
$$C^{new} = \{w_1, w_2, \dots, w_n, w_{n+1}, \dots, w_{2n}\}$$

where $ w_{i}$ with $i>n$ represents the repeated token.

Next, the pretrained language model  $LLM$ computes the attention matrix $A^{new} \in \mathbb{R}^{2n \times 2n}$ using Algorithm \ref{alg:attention-matrix-extraction}, capturing the contextual dependencies across both the original and repeated sequences.  

The backward attention mechanism is applied to strengthen semantic propagation by iteratively tracing the connections from the repeated tokens $\{w_{n+1}, w_{n+2}, \dots, w_{2n}\}$ back to the original sequence $\{w_1, w_2, \dots, w_n\}$. For a target token $w_i\in C$, the embedding $e_i$ is computed as:  

\begin{equation}\label{eq:backward_attention}
    e_i = \sum_{k=i}^{2n} \alpha'_{i,k} v_k, \text{(backward attention)}
\end{equation}

where $\alpha'_{i,k} = A^{new}_{i,k}$ is the attention weight from token $w_k $ to $w_i$, and $v_k = LLM(w_k, C^{new})$ represents the contextualized embedding of the token $w_k$.  


This approach ensures that semantic relationships, particularly those from later tokens, are comprehensively propagated and embedded into the representation. 



\begin{table*}[h!]
    \centering
    \renewcommand{\arraystretch}{1.5} 
    \setlength{\tabcolsep}{5pt} 
    \small 
    \begin{tabular}{lp{3cm}p{3cm}p{3cm}p{3cm}}
        \toprule
        \textbf{Method} & \textbf{Input Tokens}   & \multicolumn{3}{c}{\textbf{Embedding for Evaluation}} \\
        \cmidrule(lr){3-5}
        & & \textbf{Word Embedding $(w_i)$} & \textbf{Sentence Embedding (Mean Pooling)} & \textbf{Sentence Embedding (Last Pooling)} \\
        \midrule
        ReBA-$k$ (Ours)              & $\{w_1, \cdots , w_{kn}\}$          & $e_i = \sum\limits_{j=i}^{kn}\alpha'_{i,j} v_j$                 & $\frac{1}{n}\sum\limits_{j=1}^{n} e_j$                                     & $e_n$ \\
        Echo-$k$  & $\{w_1, \cdots, w_{kn}\}$        & $v_{(k-1)n+i}$                                                 & $\frac{1}{2n}\sum\limits_{j=n}^{kn} v_j$  & $v_{kn}$ \\
        Classical                  & $\{w_1, \cdots, w_n\}$           & $v_i$                                                     & $\frac{1}{n}\sum\limits_{j=1}^{n} v_j$   & $v_n$ \\
        \bottomrule
    \end{tabular}
    \caption{Introduction of our experimental settings. (1) In both ReBA and Echo method we need to repeat the original sentence,  We use $w_1,\cdots,w_n$ to denote the original input tokens, and the repeated tokens are $w_k = w_{k\%n}$ when $k>n$, $e_i$ is the new embeddings. (2) $v_i$ is the original output of $w_i$ which is denoted as $v_i:=LLM(w_i|C)$ where $C$ is the context , and $\alpha'_{i,j}$ is the $i$-th value of $j$-th column of the attention weight extracted by Algorithm \ref{alg:attention-matrix-extraction}. (3) ReBA-1 is equivalent to classical sentence embedding evaluation with last pooling strategy so we only test ReBA-1 for word embedding evaluation.}
    \label{tab:embedding_methods}
\end{table*}

\subsection{Final step: Embedding Vector Construction} 

Under different pooling strategies, the resulting sentence vectors vary. For the case of \textbf{last token pooling}, we directly use $e_n$ as the sentence encoding. In contrast, for \textbf{mean token pooling}, we take the average of all $e_i$, defined as $\frac{1}{n}\sum_{i=1}^{n} e_i$, as the sentence embedding. Notably, the order of summation can be exchanged for efficient computation:  

\begin{align}
    \frac{1}{n}\sum_{i=1}^{n}e_i &= \frac{1}{n}\sum_{i=1}^{n}\sum_{k=i}^{2n} \alpha'_{i,k} v_k \\
    & = \frac{1}{n}\sum_{k=1}^{2n}\sum_{i=1}^{\min{(n,k)}} \alpha'_{i,k} v_k \notag \\
    & = \frac{1}{n} \sum_{k=1}^{2n} \alpha'_{k} v_k, \notag
\end{align}
where $\alpha'_{k} = \sum_{i=1}^{\min{(n,k)}} \alpha'_{i,k}$ is the sum of the $k$-th column of $A^{new}$.
By exchanging the order of summation, we can first sum over each column of $ A^{\text{new}} $ and then perform the remaining calculations. This reduces the original computational complexity from $O(n^2)$ to $O(n)$.

For word embeddings, we compute the embedding $e_i$ of $w_i$ directly using Equation \ref{eq:backward_attention}, without requiring additional operations. The detailed computation process of all methods we use is shown in Table \ref{tab:embedding_methods}.

\begin{table*}[t]  
    \centering
    \resizebox{\textwidth}{!}{  
    \begin{tabular}{lccccccccc}
        \hline \hline
        Strategy & Model & Pool & Clas. & P. Cls. & Clus. & Retr. & STS & Rera. & Total Average \\
        \hline \hline
        Main results: & & & & & & & & & \\
        ReBA-2 (Ours) & GPT-2 & Last & \textbf{0.5414} & 0.5422 & \textbf{0.3243} & \textbf{0.2173} & \textbf{0.2118} & \textbf{0.2979} & \textbf{0.3634} \\
        Echo-2 & GPT-2 & Last & 0.5010 & \textbf{0.5456} & 0.2255 & 0.1757 & 0.1756 & 0.2460 & 0.2907 \\
        Classical & GPT-2 & Last & 0.4990 & 0.5413 & 0.2305 & 0.1474 & 0.1446 & 0.2101 & 0.2590 \\
        \hline 

        ReBA-2 (Ours) & LLaMA-2 & Last & \textbf{0.6889} & 0.5225 & 0.3918 & \textbf{0.6640} & 0.1949 & 0.3653 & \textbf{0.4801} \\
        Echo-2 & LLaMA-2 & Last & 0.6759 & \textbf{0.5503} & \textbf{0.4136} & 0.5629 & \textbf{0.2636} & \textbf{0.3784} & 0.4733 \\
        Classical & LLaMA-2 & Last & 0.6574 & 0.5228 & 0.3354 & 0.4667 & 0.1506 & 0.2816 & 0.3951 \\
        \hline \hline
        Ablations: Different Pool Strategy & & & & & & & & & \\
        ReBA-2 (Ours) & GPT-2 & Mean & \textbf{0.5538} & 0.5554 & \textbf{0.3275} & \textbf{0.4484} & \textbf{0.2567} & \textbf{0.3000} & \textbf{0.3977} \\
        Echo-2 & GPT-2 & Mean & 0.5480 & \textbf{0.5588} & \textbf{0.3275} & 0.3691 & 0.2360 & 0.2964 & 0.3734 \\
        Classical & GPT-2 & Mean & 0.5465 & 0.5452 & 0.3166 & 0.2959 & 0.2515 & 0.2959 & 0.3499 \\

        \hline
        Ablations: More Repetition & & & & & & & & & \\
        Echo-3 & GPT-2 & Last & 0.4973 & 0.5473 & 0.1969 & 0.1845 & 0.1845 & 0.2488 & 0.2924 \\
        ReBA-3 (Ours) & GPT-2 & Last & 0.5422 & 0.5446 & 0.3246 & 0.2119 & 0.2284 & 0.3010 & 0.3699 \\

        \hline \hline
        Comparison: Bidirectional model & & & & & & & & & \\
        Classical & BERT & Mean & 0.6734 & 0.5435 & 0.4225 & 0.5537 & 0.2594 & 0.3445 & 0.4658 \\
        Classical & BERT & CLS & 0.6628 & 0.5604 & 0.3598 & 0.3003 & 0.2070 & 0.2546 & 0.3679 \\


        \hline \hline
        
    \end{tabular}
    }
    \caption{Zero-shot average scores on C-MTEB to evaluate the performance of \emph{Sentence Embeddings}. The top two rows are the main results, testing the scores of ReBA, simple repetition and traditional encoding on GPT-2 and LLaMA-2-7B. The third row is an ablation experiment, testing different pooling strategies. The fourth row is an ablation experiment with more repetitions, testing the effect of repeating twice. The fifth row is a comparison experiment with bidirectional model using mean pooling or the first token 'CLS', BERT refers to bert-base-chinese model  with 102M params here.}
    \label{tab:main_results}
\end{table*}

\begin{algorithm}
    \caption{Word Embedding with ReBA Mechanism}
    \textbf{Input:} Text sequence $C=\{w_1, w_2, \dots, w_n\}$, pretrained language model $LLM$, target word $w_i$.\\
    \textbf{Output:} Word embedding $e_i$.\\
    1. Duplicate the input: $C^{new}= \{w_1, w_2, \dots, w_{2n}\}$ as new input.\\
    2. Compute the attention matrix $A^{new}\in\mathbb{R}^{2n\times 2n}$ using Algorithm \ref{alg:attention-matrix-extraction}.\\
    3. Extract the embedding $e_i$ of $w_i$ \{$i\le n$\} based on $A^{new}$: $e_i = \sum^{2n}_{k=i}\alpha'_{i,k}v_k$ where $\alpha'_{i,k}=A^{new}_{i,k}$ is the $i$-th value of $k$-th column of $A^{new}$ and $v_k = LLM(w_k|C)$.\\
    5. \textbf{Return} $e_i$.
    \label{alg:word_embedding_construction}
\end{algorithm}

\begin{algorithm}[H]
    \caption{Sentence Embedding with ReBA Mechanism}
    \textbf{Input:} Text sequence $C = \{w_1, w_2, \dots, w_n\}$, pretrained language model $LLM$.\\
    \textbf{Output:} Sentence embedding $E$.
        
    1. Extract word embeddings $e_i$ for each token $w_i$ in $C$ using Algorithm \ref{alg:word_embedding_construction}.

    2. \textbf{If} last token pool: $E = e_n$.\\
    \hspace{4em} \textbf{Else if} mean token pool: $E = \frac{1}{n} \sum_{i=1}^{n} e_i$.

    3. \textbf{Return} $E$.
    \label{alg:embedding-construction}
\end{algorithm}

\subsection{Time efficiency}\label{sec:analysis}


Assuming the model has \( I \) hidden layers, each with \( J \) attention heads, and an input sequence of length \( n \), the traditional method has a time complexity of \( O(IJn^2) \). This arises from extracting \( I \times J \) attention matrices, each of size \( n \times n \), and performing operations like symmetry computation and maximum attention update, both of which take \( O(n^2) \) for a single matrix.


\section{Experiments}
\subsection{Datasets}
We will test our method on several datasets, including text classification, text clustering, text pair classification, text reranking, text retrieval, sentence similarity, etc. In particular, we will also test its performance on Chinese polysemous word understanding task.

\subsubsection{Chinese Massive Text Embedding Benchmark(For sentence embedding)}
To evaluate sentence embedding, we use the Chinese Massive Text Embedding Benchmark (C-MTEB), it is a collection of datasets from six categories: classification, clustering, pair classification, reranking, retrieval, sentence similarity and contains 31 datasets in total. Due to the scale of this dataset, we will only test two unidirectional language models: the fine-tuned versions of GPT-2 and LLaMA-2-7B's Chinese pre-trained models.

\subsubsection{Chinese Polysemous Word Disambiguation Dataset (For word embedding)}
These datasets evaluate a model's ability to understand polysemous words and assess word embedding performance. The first dataset, the Sentence Level Polysemous Words Classification (SLPWC) subset of the \href{https://github.com/FlagOpen/FlagEval/tree/master/csem}{C-SEM}\footnote{Accessible at: \url{https://github.com/FlagOpen/FlagEval/tree/master/csem}} (Chinese Semantic Evaluation Dataset) benchmark, contains 300 questions where the task is to identify which option represents a different meaning of a polysemous word. We extract embeddings of the target word in different contexts using an LLM, and the correct answer is determined by summing the embedding distances (Cosine or Euclidean). The second dataset, Word Sense Disambiguation (WSD, \cite{yan2023construction}), is a Chinese semantic dataset used for word sense disambiguation tasks. We evaluate performance based on accuracy and adapt it into a 4-choice format to suit our algorithm.


\subsection{Model}

In our experiments, we use pretrained autoregressive models based on the Transformers architecture, including fine-tuned versions of GPT-2 and LLaMA-2's Chinese pretrained models GPT-2-Chinese, LLaMA-2-Chinese-7B, as well as models like Qwen-7B, BaiChuan-7B, and Falcon-7B, which are suitable for evaluation on Chinese dataset. All model details are presented in Appendix \ref{sec_appendix:model_detail}.


        
        

\subsection{Results}

\begin{figure*}[h!]
    \centering
    \includegraphics[width=0.8\textwidth]{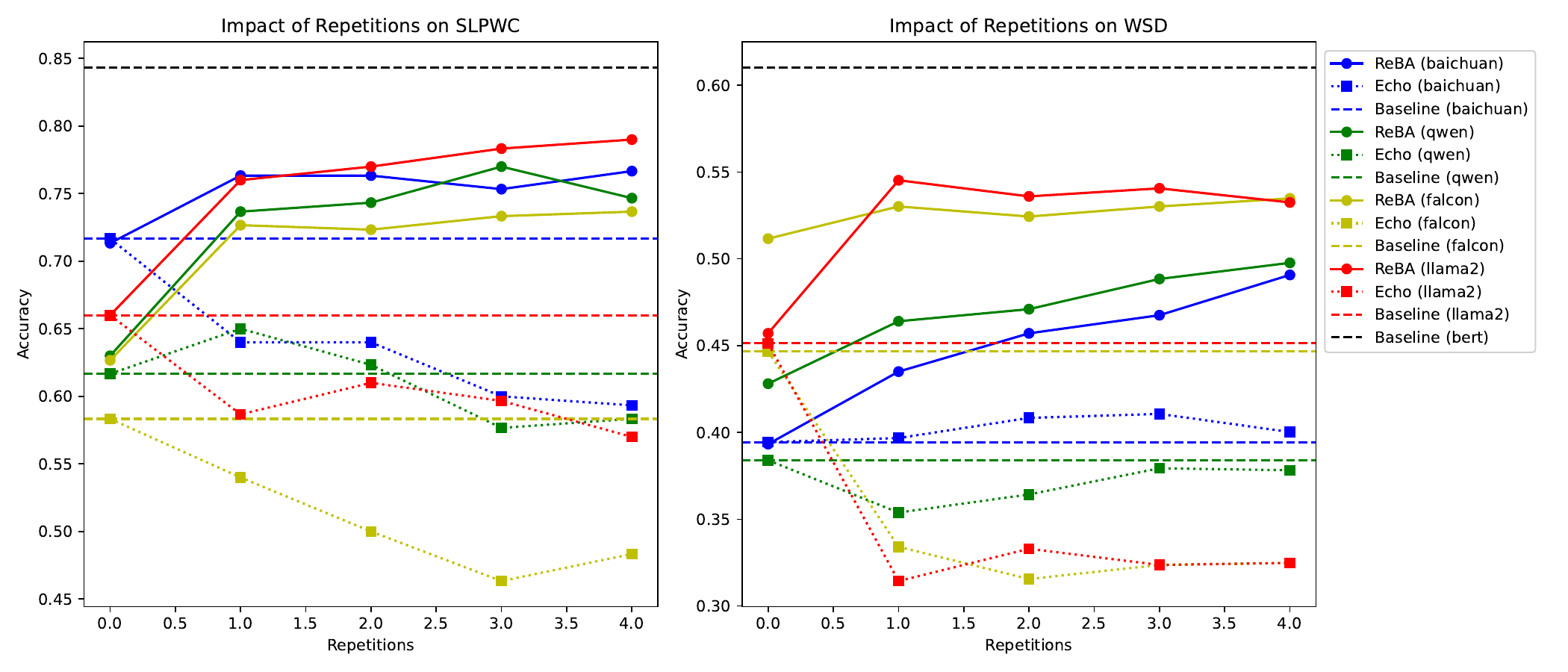}
    \caption{Performance on SLPWC and WSD tasks using Euclidean distances to evaluate \emph{word embeddings}. The results show that ReBA encoding significantly enhances model performance on polysemous word understanding tasks. While performance fluctuates with the number of repetitions, increasing the repetition count does not necessarily lead to significant improvements. Based on this experiment, we observe that simple sentence repetition is not effective for improving word-level embeddings and only contributes to sentence-level understanding. Furthermore, the backward attention mechanism remains crucial for achieving further performance enhancements.}
    \label{fig:word_results}
\end{figure*}

The overall results (see Table \ref{tab:main_results} and Figure \ref{fig:word_results}) show that the ReBA encoding method significantly outperforms traditional encoding methods for both sentence and word embeddings. In particular, our method demonstrates a significant improvement in accuracy on semantic understanding tasks. This indicates that our algorithm effectively enhances language models' understanding of natural language in zero-shot setting.

\subsubsection{Sentence embedding evaluation}


        

We primarily evaluated GPT-2-Chinese(we call it GPT-2 for simplicity) and LLaMA-2-Chinese-7B(LLaMA-2, for simplicity) on C-MTEB, comparing the performance of our method with traditional encoding approaches under different pooling strategies. Across nearly all tasks, our method demonstrated significant improvements over traditional methods. The results can be concluded as follows:

    (1). \emph{ReBA consistently enhances unidirectional language models}, achieving greater improvements on GPT-2 compared to simple repetition. For LLaMA-2, while our method outperforms traditional encoding, the gains on certain tasks (e.g., STS) are less pronounced compared to simple repetition.  

    (2). \emph{Increasing the repetition count does not yield additional benefits.} We compared the effects of repeating once and twice (ReBA-2 and ReBA-3), finding that while repeating twice achieves better results, the improvement is marginal. This conclusion holds for both ReBA and simple repetition.  

    (3). \emph{Our method remains effective across different pooling strategies.} With last-pooling, the algorithm achieves substantial improvements, while the gains with mean-pooling are comparatively smaller. 


\subsubsection{Word embedding evaluation}
We evaluate the performance of LLaMA-2-Chinese-7B, Qwen-7B, BaiChuan-7B, and Chinese-Falcon-7B and Bert-base-chinese (102M) on the Chinese polysemous word understanding task using the Chinese SEMantic evaluation dataset (C-SEM) and the Word Sense Disambiguation dataset (WSD). The results are presented in Figure \ref{fig:word_results}:

(1) \emph{ReBA encoding performs exceptionally well as a word embedding method}, and ReBA significantly surpassing classical embeddings.  

(2) \emph{Backward attention is the essential operation}: We tested Echo embeddings as word embeddings by using the embedding of the target word's last occurrence. Surprisingly, repetition alone degraded performance, but adding backward attention significantly improved it. ReBA embeddings outperformed both Echo and classical embeddings by a wide margin.

(3) \emph{Increasing the repetition count does not yield additional
benefits}. Also in the word embedding task, we find that repeating twice does not bring more benefits than repeating once, this conclusion is consistent with the sentence embedding task.

\section{Conclusion}
We propose a context-enhanced encoding method based on backward attention mechanisms and text repetition. Experimental results demonstrate significant improvements on LLMs across multiple tasks. On sentence embedding evaluation datasets, we observed that the backward attention mechanism was not a decisive factor—simply repeating sentences was sufficient to improve sentence vector quality. This effect was particularly pronounced in larger models like LLaMA-2, where the gains from backward attention were minimal.

However, on word embedding evaluation datasets, the backward attention mechanism played a crucial role. In these cases, simply repeating sentences often led to performance degradation, whereas incorporating backward attention resulted in substantial improvements. Consequently, we developed a more general method for improving language model encoding quality. 

\section{Discussion}

In this work, we analyzed the computational overhead of processing repeated text in large language models (LLMs). Repeating the entire text doubles the sequence length, introducing an additional computational overhead of \( O(L^2) \), which becomes prohibitively expensive for long inputs. We propose a potential alternative for future study:

We first encode the original sequence \( S \) of length \( L \) using an LLM to obtain the initial embeddings. Next, we divide \( S \) into subsequences \( S_0, S_1, \dots, S_n \), each of length \( L_0 \). For each subsequence \( S_i \), we repeat it to form \( 2S_i \) and input this repeated sequence into the LLM to obtain the updated encoding of \( S_i \).  

Using the attention weights between the original and repeated embeddings of \( S_i \), we apply backward attention to enhance the initial embeddings of \( S_i \), resulting in the refined ReBA embeddings. This approach reduces the additional computational overhead caused by repetition from \( O(L^2) \) to \( O(L \cdot L_0) \). When \( L_0 \) is sufficiently small, the additional cost becomes negligible.

\bibliography{refs.bib}

\bibliographystyle{icml2025}


\newpage
\appendix
\section{Appendix}\label{sec:appendix}
\subsection{Datasets}
\subsubsection{Sentence embedding evaluation}

To evaluate sentence embeddings, we use the Chinese Massive Text Embedding Benchmark (C-MTEB), which is a collection of datasets across six categories: classification, clustering, pair classification, reranking, retrieval, and sentence similarity. In total, there are 31 datasets. The dataset is available at https://huggingface.co/datasets/C-MTEB, and the leaderboard can be found at https://huggingface.co/spaces/mteb/leaderboard.

\textbf{Data preprocessing:}

A brief overview of the C-MTEB datasets is provided below:

\begin{figure}[!h]
    \centering
    \includegraphics[width=0.45\textwidth]{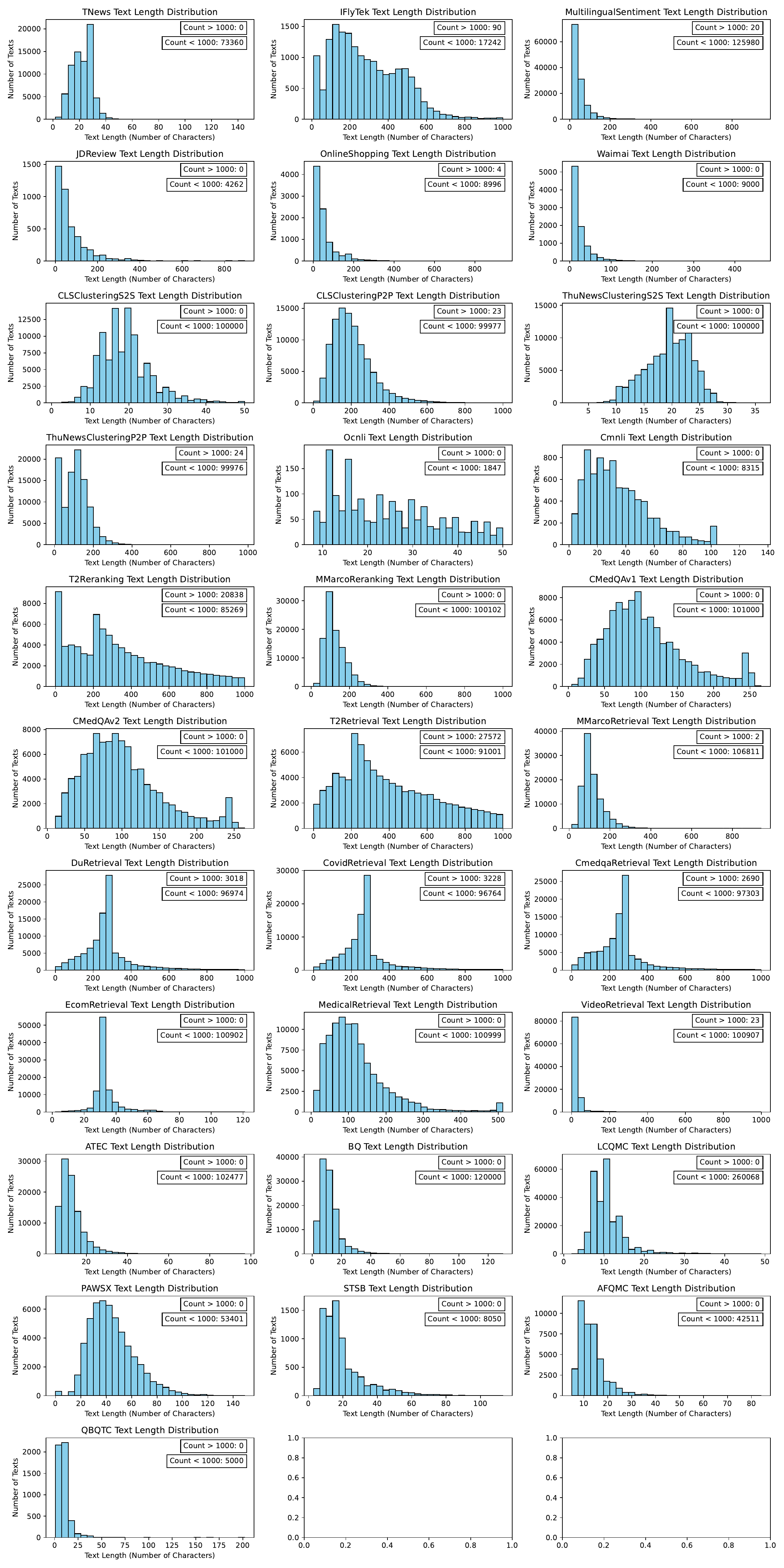}
    \caption{Information about C-MTEB, with most text lengths within 1000 tokens.}
    \label{fig:c-metb}
\end{figure}

In our experiments, since GPT models have an input sequence limit of 512 tokens, we applied text truncation accordingly. By analyzing the text length distribution in the C-MTEB dataset, we found that most texts are under 512 tokens (Figure \ref{fig:c-metb}), and a substantial portion remains below 1024 tokens even when repeated three times. Therefore, for our experiments with LLaMA-2, in order to avoid potential memory overflow issues caused by a small number of long texts, we truncated the text to a maximum length of 1024 tokens.

The detailed results can be found in Table \ref{tab:main_results}. Additionally, Table \ref{tab:appendix_gpt_last_scores} presents the results for GPT-2 on C-MTEB using the last pooling strategies.

\begin{table*}[htbp]
    \centering
    \scriptsize
    \begin{tabular}{@{}llccccc@{}}
    \toprule
    \textbf{Task Type} & \textbf{Task Name} & \textbf{Classical} & \textbf{Repetition only (1 time)} & \textbf{Repetition only (2 times)} & \textbf{ReBA} & \textbf{ReBA-3} \\ \midrule
    Classification       & TNews                  & 0.2775        & 0.2679        & 0.2766        & 0.3061        & 0.3055        \\ 
    Classification       & IFlyTek                & 0.2288        & 0.1915        & 0.2099        & 0.3094        & 0.3099        \\ 
    Classification       & MultilingualSentiment  & 0.4851        & 0.4838        & 0.4865        & 0.5044        & 0.5064        \\ 
    Classification       & JDReview               & 0.6989        & 0.7092        & 0.7069        & 0.7357        & 0.7342        \\ 
    Classification       & OnlineShopping         & 0.6637        & 0.6626        & 0.6681        & 0.6943        & 0.6948        \\ 
    Classification       & Waimai                 & 0.6398        & 0.6687        & 0.6580        & 0.6991        & 0.7026        \\ \midrule
    Clustering           & CLSClusteringS2S       & 0.1282        & 0.1505        & 0.1446        & 0.2289        & 0.2308        \\ 
    Clustering           & CLSClusteringP2P       & 0.2086        & 0.1077        & 0.1480        & 0.3081        & 0.3107        \\ 
    Clustering           & ThuNewsClusteringS2S   & 0.2419        & 0.2587        & 0.2678        & 0.3250        & 0.3246        \\ 
    Clustering           & ThuNewsClusteringP2P   & 0.3433        & 0.2708        & 0.3417        & 0.4352        & 0.4322        \\ \midrule
    Pair Classification  & Ocnli                  & 0.5338        & 0.5306        & 0.5349        & 0.5382        & 0.5382        \\ 
    Pair Classification  & Cmnli                  & 0.5488        & 0.5639        & 0.5562        & 0.5462        & 0.5511        \\ \midrule
    Reranking            & T2Reranking            & 0.5254        & 0.5539        & 0.5462        & 0.5570        & 0.5613        \\ 
    Reranking            & MMarcoReranking        & 0.0263        & 0.0574        & 0.0535        & 0.0709        & 0.0714        \\ 
    Reranking            & CMedQAv1               & 0.1399        & 0.1800        & 0.1852        & 0.2788        & 0.2811        \\ 
    Reranking            & CMedQAv2               & 0.1488        & 0.2037        & 0.1993        & 0.2852        & 0.2903        \\ \midrule
    Retrieval            & T2Retrieval            & 0.0564        & 0.1018        & 0.0972        & 0.2079        & 0.2265        \\ 
    Retrieval            & MMarcoRetrieval        & 0.2201        & 0.3305        & 0.3316        & 0.4766        & 0.4925        \\ 
    Retrieval            & DuRetrieval            & 0.1076        & 0.1456        & 0.1512        & 0.3076        & 0.3417        \\ 
    Retrieval            & CovidRetrieval         & 0.0464        & 0.0105        & 0.0200        & 0.1628        & 0.1786        \\ 
    Retrieval            & CmedqaRetrieval        & 0.2199        & 0.3171        & 0.3105        & 0.4196        & 0.4188        \\ 
    Retrieval            & EcomRetrieval          & 0.2430        & 0.3880        & 0.3660        & 0.5330        & 0.5620        \\ 
    Retrieval            & MedicalRetrieval       & 0.0820        & 0.1700        & 0.1520        & 0.2860        & 0.2990        \\ 
    Retrieval            & VideoRetrieval         & 0.2040        & 0.4490        & 0.3710        & 0.5670        & 0.5810        \\ \midrule
    STS                  & ATEC                   & 0.1321        & 0.1476        & 0.1453        & 0.1652        & 0.1769        \\ 
    STS                  & BQ                     & 0.1902        & 0.2239        & 0.2134        & 0.2532        & 0.2561        \\ 
    STS                  & LCQMC                  & 0.1116        & 0.2111        & 0.1830        & 0.3199        & 0.3329        \\ 
    STS                  & PAWSX                  & 0.1234        & 0.1185        & 0.1195        & 0.1331        & 0.1327        \\ 
    STS                  & STSB                   & 0.2563        & 0.3429        & 0.3268        & 0.3469        & 0.3660        \\ 
    STS                  & AFQMC                  & 0.0798        & 0.0780        & 0.0789        & 0.0997        & 0.1060        \\ 
    STS                  & QBQTC                  & 0.1189        & 0.1696        & 0.1628        & 0.1651        & 0.1512        \\ \midrule
    Total average       & N/A                    & 0.2591        & 0.2924        & 0.2907        & 0.3634        & 0.3699        \\ \bottomrule
    \end{tabular}
    \caption{Main Results: Zero-shot scores of GPT-2 models on C-MTEB under last pooling strategy with different methods. 'Classical' refers to the traditional encoding method, 'Repetition only (1 time)' and 'Repetition only (2 times)' refer to the methods that only repeat the text without backward attention, and 'ReBA' and 'ReBA-3' refer to our proposed method with one and two repetitions, respectively, the scores we choose are 'accuracy', 'v\_measure'
    , 'map', 'cos\_sim :accuracy', 'cos\_sim :pearson', 'recall\_at\_1000' for Classification, Clustering, Reranking, Pair Classification, STS, Retrieval.}
    \label{tab:appendix_gpt_last_scores}
\end{table*}

\begin{table*}[h]
    \centering
    \begin{tabular}{l l c c c}
    \toprule
    \textbf{Task Type} & \textbf{Task Name} & \textbf{Classical} & \textbf{Echo-2} & \textbf{ReBA-2} \\
    \midrule
    Classification & TNews & 0.3048 & 0.3090 & 0.3102 \\
    Classification & IFlyTek & 0.3414 & 0.3179 & 0.3474 \\
    Classification & MultilingualSentiment & 0.4952 & 0.5062 & 0.5058 \\
    Classification & JDReview & 0.7272 & 0.7432 & 0.7373 \\
    Classification & OnlineShopping & 0.6957 & 0.7021 & 0.7028 \\
    Classification & Waimai & 0.7150 & 0.7095 & 0.7193 \\
    \midrule
    Clustering & CLSClusteringS2S & 0.2188 & 0.2473 & 0.2430 \\
    Clustering & CLSClusteringP2P & 0.3186 & 0.3189 & 0.3208 \\
    Clustering & ThuNewsClusteringS2S & 0.3065 & 0.3222 & 0.3217 \\
    Clustering & ThuNewsClusteringP2P & 0.4225 & 0.4214 & 0.4245 \\
    \midrule
    Pair Classification & Ocnli & 0.5452 & 0.5501 & 0.5485 \\
    Pair Classification & Cmnli & 0.5452 & 0.5675 & 0.5624 \\
    \midrule
    Reranking & T2Reranking & 0.5553 & 0.5518 & 0.6061 \\
    Reranking & MMarcoReranking & 0.0365 & 0.0629 & 0.0431 \\
    Reranking & CMedQAv1 & 0.2326 & 0.2771 & 0.2680 \\
    Reranking & CMedQAv2 & 0.2587 & 0.2941 & 0.2833 \\
    \midrule
    Retrieval & T2Retrieval & 0.1360 & 0.1992 & 0.3603 \\
    Retrieval & MMarcoRetrieval & 0.2843 & 0.4798 & 0.4190 \\
    Retrieval & DuRetrieval & 0.1698 & 0.2856 & 0.3261 \\
    Retrieval & CovidRetrieval & 0.3335 & 0.1581 & 0.7758 \\
    Retrieval & CmedqaRetrieval & 0.3590 & 0.4296 & 0.3962 \\
    Retrieval & EcomRetrieval & 0.3860 & 0.5550 & 0.4970 \\
    Retrieval & MedicalRetrieval & 0.1660 & 0.2680 & 0.2360 \\
    Retrieval & VideoRetrieval & 0.5330 & 0.5780 & 0.5770 \\
    \midrule
    STS & ATEC & 0.1678 & 0.1950 & 0.1886 \\
    STS & BQ & 0.2918 & 0.2761 & 0.2926 \\
    STS & LCQMC & 0.4296 & 0.3853 & 0.4352 \\
    STS & PAWSX & 0.1448 & 0.1036 & 0.1214 \\
    STS & STSB & 0.4450 & 0.5140 & 0.4917 \\
    STS & AFQMC & 0.1018 & 0.1117 & 0.1105 \\
    STS & QBQTC & 0.1799 & 0.1366 & 0.1571 \\
    \midrule
    Total average & N/A & 0.3499 & 0.3734 & 0.3977 \\
    \bottomrule
    \end{tabular}
    \caption{Ablations: Zero-shot scores of GPT-2 models on C-MTEB under mean pooling strategy with different methods. 'Classical' refers to the traditional encoding method, 'Echo-2' refer to the methods that only repeat the text without backward attention, and 'ReBA' refer to our proposed method with one repetition, respectively, the scores we choose are 'accuracy', 'v\_measure'
    , 'map', 'cos\_sim', 'pearson', 'recall\_at\_1000'.}
    \label{tab:appendix_gpt_mean_scores}
\end{table*}

\begin{table*}[h]
    \centering
    \begin{tabular}{l l c c c c}
    \toprule
    \textbf{Task Type} & \textbf{Task Name} & \textbf{Classical} & \textbf{Echo-2} & \textbf{Echo-3} & \textbf{ReBA-2} \\
    \midrule
    Classification & TNews & 0.5194 & 0.5264 & 0.5203 & 0.5334 \\
    Classification & IFlyTek & 0.3663 & 0.4082 & 0.3552 & 0.4429 \\
    Classification & MultilingualSentiment & 0.6474 & 0.6554 & 0.6476 & 0.6645 \\
    Classification & JDReview & 0.7645 & 0.7717 & 0.7497 & 0.7976 \\
    Classification & OnlineShopping & 0.8521 & 0.8724 & 0.8674 & 0.8745 \\
    Classification & Waimai & 0.7951 & 0.8214 & 0.8226 & 0.8206 \\
    \midrule
    Clustering & CLSClusteringS2S & 0.2446 & 0.2836 & 0.3000 & 0.3021 \\
    Clustering & CLSClusteringP2P & 0.2858 & 0.3169 & 0.3049 & 0.2972 \\
    Clustering & ThuNewsClusteringS2S & 0.4841 & 0.5572 & 0.5522 & 0.5702 \\
    Clustering & ThuNewsClusteringP2P & 0.3272 & 0.4968 & 0.5216 & 0.3976 \\
    \midrule
    Pair Classification & Ocnli & 0.5181 & 0.5463 & 0.5355 & 0.5176 \\
    Pair Classification & Cmnli & 0.5276 & 0.5543 & 0.5498 & 0.5275 \\
    \midrule
    Reranking & T2Reranking & 0.5949 & 0.5806 & 0.5708 & 0.6185 \\
    Reranking & MMarcoReranking & 0.0437 & 0.0774 & 0.0867 & 0.0619 \\
    Reranking & CMedQAv1 & 0.2390 & 0.3976 & 0.3948 & 0.3667 \\
    Reranking & CMedQAv2 & 0.2489 & 0.4581 & 0.4713 & 0.4142 \\
    \midrule
    Retrieval & T2Retrieval & 0.4387 & 0.4699 & 0.3840 & 0.6334 \\
    Retrieval & MMarcoRetrieval & 0.6340 & 0.7425 & 0.7423 & 0.7923 \\
    Retrieval & DuRetrieval & 0.5701 & 0.7322 & 0.6740 & 0.7936 \\
    Retrieval & CovidRetrieval & 0.5896 & 0.2819 & 0.1923 & 0.6723 \\
    Retrieval & CmedqaRetrieval & 0.3713 & 0.6642 & 0.6775 & 0.5355 \\
    Retrieval & EcomRetrieval & 0.5990 & 0.7180 & 0.7700 & 0.8310 \\
    Retrieval & MedicalRetrieval & 0.2090 & 0.4760 & 0.5030 & 0.4170 \\
    Retrieval & VideoRetrieval & 0.3220 & 0.4190 & 0.5780 & 0.6370 \\
    \midrule
    STS & ATEC & 0.1328 & 0.1880 & 0.1793 & 0.1876 \\
    STS & BQ & 0.1852 & 0.3185 & 0.3199 & 0.2385 \\
    STS & LCQMC & 0.2397 & 0.4924 & 0.4886 & 0.3136 \\
    STS & PAWSX & 0.1113 & 0.1402 & 0.1399 & 0.1112 \\
    STS & STSB & 0.2656 & 0.4559 & 0.4187 & 0.3443 \\
    STS & AFQMC & 0.1095 & 0.1549 & 0.1373 & 0.1488 \\
    STS & QBQTC & 0.0105 & 0.0956 & 0.1907 & 0.0206 \\
    \midrule
    Total average & N/A & 0.3951 & 0.4733 & 0.4724 & 0.4801 \\
    \bottomrule
    \end{tabular}
    \caption{Ablations: Zero-shot scores of LLaMA-2 model on C-MTEB under last pooling strategy with different methods. 'Classical' refers to the traditional encoding method, 'Echo-2' refer to the methods that only repeat the text without backward attention, and 'ReBA-2' refer to our proposed method with one repetition, respectively, the scores we choose are 'accuracy', 'v\_measure'
    , 'map', 'cos\_sim', 'pearson', 'recall\_at\_1000'.}
    \label{tab:appendix_llama_last_scores}
\end{table*}

\subsubsection{Word embedding evaluation}\label{appendix:word_embedding_evaluation}
To  evaluate  word  embeddings,  we  use  the  Chinese  SEMantic  evaluation  dataset  (C-SEM),  a  benchmark  dataset  for  semantic  evaluation.  We  use  the  Sentence  Level  Polysemous  Words  Classification  (SLPWC)  subset  of  C-SEM  as  our  evaluation  dataset.  This  subset  is  designed  to  test  a  model’s  ability  to  understand  polysemy  (i.e.,  words  with  multiple  meanings).  The  evaluation  involves  presenting  a  word  in  different  contexts  and  expecting  the  model  to  identify  semantic  differences.

1. \textbf{SLPWC}: 
The SLPWC dataset contains 300 polysemous words, each of which appears in four sentences. In three of the sentences, the polysemous word has the same meaning, while in the remaining sentence, the word has a different meaning. The task is to identify the sentence with the different meaning, the data presents a question format: `Which of the following sentences uses `word' differently from the others? A. sentence1; B. sentence2; C. sentence3; D. sentence4.’ An example from the dataset is presented in section \ref{appendix:word_embedding_evaluation}.

2. \textbf{WSD}:

The Word Sense Disambiguation (WSD) dataset contains 1,023 polysemous words, each associated with multiple meanings, and each meaning linked to several example sentences. The dataset is structured as: \{word: \{sense1: [sentence1, sentence2, sentence3]; sense2: [sentence4]\}\}. To ensure consistent evaluation, we converted the WSD dataset into the SLPWC format. Specifically, three sentences are randomly selected from one meaning and one from another. The transformed data presents a question format: `Which of the following sentences uses `word' differently from the others? A. sentence1; B. sentence2; C. sentence3; D. sentence4.’ Below is an example of question in these two dataset:

\begin{CJK}{UTF8}{gbsn}
    
Question：以下哪句话中“中学”的意思(或用法)与其他句子不同。

A. \textbf{中学}教育在塑造青少年的品德、知识和技能方面起着重要的作用。

B. 曾纪泽、张自牧、郑观应、陈炽、薛福成等大抵讲“\textbf{中学}为体，西学为用”的人，无不持“西学中源”说。

C. \textbf{中学}是为了培养青少年的综合素质而设立的教育机构。

D. 我们的学校是一所提供\textbf{中学}教育的优秀学校，致力于为学生提供高质量的教育和培养。

it ask which of the following sentences is the meaning (or usage) of "中学" different from the other sentences, the correct answer 'B'(means Chinese culture, "中学" in A,C,D means 'middle school' ) for this problem will also show in the dataset as label.
\end{CJK}

            
            
            




\subsubsection{Attention Matrix Processing}

We also conducted comparative experiments with symmetric attention matrices and last-layer-only attention. The results are shown in Table \ref{tab:main_results}. Both methods underperformed compared to our approach.

\subsection{Model detail}\label{sec_appendix:model_detail}

Here we provide some details of the models we use in our experiments: \href{https://huggingface.co/google-bert/bert-base-chinese}{BERT-base-chinese}\footnote{BERT-base-chinese:\url{https://huggingface.co/google-bert/bert-base-chinese}}, \href{https://huggingface.co/ckiplab/gpt2-base-chinese}{GPT2-base-chinese}\footnote{GPT2-base-chinese:\url{https://huggingface.co/ckiplab/gpt2-base-chinese}} ,\href{https://huggingface.co/LinkSoul/Chinese-Llama-2-7b}{Chinese-llama-2-7b}\footnote{Chinese-llama-2-7b:\url{https://huggingface.co/LinkSoul/Chinese-Llama-2-7b}} , \href{https://huggingface.co/Qwen/Qwen-7B}{Qwen-7B}\footnote{Qwen-7B:\url{https://huggingface.co/Qwen/Qwen-7B}}, \href{https://huggingface.co/Linly-AI/Chinese-Falcon-7B}{Chinese-Falcon-7b}\footnote{Falcon-7B: \url{https://huggingface.co/Linly-AI/Chinese-Falcon-7B}} and \href{https://huggingface.co/baichuan-inc/Baichuan-7B}{BaiChuan-7B}\footnote{BaiChuan-7B:\url{https://huggingface.co/baichuan-inc/Baichuan-7B}} as our models.

The BERT-base-chinese model, developed by Google, is a pre-trained language model tailored for Chinese natural language processing tasks. Built on the BERT architecture, it comprises 12 layers, 768 hidden units, and 12 attention heads, totaling approximately 110 million parameters. Pre-trained on large Chinese corpora, including Chinese Wikipedia, using Masked Language Modeling (MLM) and Next Sentence Prediction (NSP) objectives, it effectively captures word and sentence-level semantics. This model serves as a robust baseline for tasks such as text classification, named entity recognition, and question answering, offering strong performance across diverse Chinese NLP applications.

The GPT series is a family of pretrained models based on the Transformer architecture, with GPT-2\cite{radford2019language} being the second-generation generative pretrained model released by OpenAI in 2019. We used the Chinese version of GPT-2, GPT2-base-chinese, which is fine-tuned on Traditional Chinese datasets to better adapt to Chinese contexts. It has 12 layers, 768 hidden units, and 12 attention heads.

LLaMA-2 (Large Language Model Meta AI 2) \cite{touvron2023llama} is the second-generation large language model released by Meta (formerly Facebook), designed to handle various language tasks, including text generation, comprehension, and question-answering. It is an enhanced version of the original LLaMA model, featuring improved performance and adaptability. We used a Simplified Chinese fine-tuned version of LLaMA-2 for our experiments. It has 32 layers, 4096 hidden units, and 32 attention heads.

Qwen-7B (Tongyi Qianwen) is a unidirectional language model developed by Alibaba Group. With 7 billion parameters, it is designed to handle a wide range of tasks, including text generation, content summarization, and intelligent decision-making. The model excels in Chinese language processing and supports multilingual tasks, making it suitable for diverse real-world applications.

Baichuan-7B is a unidirectional language model with 7 billion parameters, developed in China for Chinese and multilingual NLP tasks. It demonstrates strong capabilities in machine translation, text classification, and semantic understanding. The model is widely recognized for its adaptability and practical application across various industries.

The Chinese-Falcon-7B, developed by Linly-AI, is an adaptation of the original Falcon architecture, tailored specifically for Chinese natural language processing tasks. With 32 Transformer layers, 71 attention heads per layer, and a hidden size of 4544, it retains the efficient design of Falcon while being pre-trained on a large-scale Chinese corpus. This specialization enables superior performance in Chinese text understanding and generation, making it suitable for applications such as summarization, sentiment analysis, and conversational AI.

Here are the basic statistics of the models used in our experiments:

\begin{table}[H]
    \centering
    \begin{tabular}{lcccc}
        \hline \hline
        Model & Layers & Hidden Units & Attention Heads \\
        \hline
        BERT & 12 & 768 & 12 \\
        GPT-2 & 12 & 768 & 12 \\
        LLaMA-2 & 32 & 4096 & 32 \\
        Qwen & 32 & 4096 & 32 \\
        BaiChuan & 32 & 4096 & 32 \\
        Falcon & 32 & 4544 & 71 \\
        \hline \hline
    \end{tabular}
    \caption{Basic Statistics of the Models}
    \label{tab:model_statistics}
\end{table}

\subsection{Details of the sentence evaluation: task description and metrics}\label{sec_appendix:task_description}
There are six types of tasks in the C-MTEB dataset: Classification, Clustering, Pair Classification, Reranking, Retrieval, and STS. Each task has specific evaluation metrics and requirements, as detailed below:

The \textbf{Classification} task involves assigning labels to text inputs from predefined categories. For example, the TNews dataset requires predicting news categories based on headlines. The primary evaluation metric for this task is accuracy, defined as:  
$$
\text{Accuracy} = \frac{N_c}{N}
$$  
where $N_c$ is the number of correct predictions, and $N$ is the total number of samples.

The \textbf{Clustering} task groups text samples based on their semantic similarity without predefined labels. An example is the CLSClusteringS2S dataset, where similar sentences need to be grouped together. The evaluation metric is V-measure, defined as:  
$$
V = 2 \times \frac{H \times C}{H + C}
$$  
where $H$ represents homogeneity, and $C$ represents completeness.

The \textbf{Reranking} task focuses on reordering retrieved documents by their relevance to a query. For instance, in the T2Reranking dataset, the task involves ranking candidate documents for search queries. The main evaluation metric is Mean Average Precision (MAP). For a query $q$, the Average Precision (AP) is defined as:  
$$
AP_q = \frac{1}{R_q} \sum_{k=1}^{n} P(k) \cdot \delta(k)
$$  
where $R_q$ is the number of relevant documents for query $q$, $P(k)$ is the precision at position $k$, and $\delta(k)$ is an indicator function that equals 1 if the document at position $k$ is relevant, otherwise 0. MAP is the mean of AP over all queries.

The \textbf{Pair Classification} task determines whether two sentences are semantically equivalent. An example dataset is Ocnli, which focuses on classifying sentence pairs into categories such as entailment, contradiction, or neutral. The evaluation metrics used are cosine similarity-based accuracy and Pearson correlation. Cosine similarity between embedding vectors $\mathbf{u}$ and $\mathbf{v}$ is defined as:  
$$
\cos(\mathbf{u}, \mathbf{v}) = \frac{\mathbf{u} \cdot \mathbf{v}}{\|\mathbf{u}\| \|\mathbf{v}\|}
$$  
Cosine similarity-based accuracy measures the alignment between predicted similarity and semantic equivalence, while Pearson correlation evaluates the linear relationship between cosine similarity scores and human-labeled ground truth.

The \textbf{STS (Semantic Textual Similarity)} task evaluates the degree of semantic similarity between pairs of sentences by comparing their embeddings. For example, the ATEC dataset assesses sentence similarity in financial question matching scenarios. The primary evaluation metric used is Pearson correlation, which quantifies the linear relationship between the predicted similarities and the ground truth labels.
The Pearson correlation coefficient between predicted cosine similarities $\hat{y}$ and true similarities $y$ is computed as:
$$
r = \frac{\sum_{i=1}^{n} (\hat{y}_i - \bar{\hat{y}})(y_i - \bar{y})}{\sqrt{\sum_{i=1}^{n} (\hat{y}_i - \bar{\hat{y}})^2} \sqrt{\sum_{i=1}^{n} (y_i - \bar{y})^2}}
$$

The \textbf{Retrieval} task evaluates a model’s ability to identify relevant documents in a large collection. For example, the MMarcoRetrieval dataset involves retrieving relevant documents for search queries. The primary evaluation metric is Recall at 1000, defined as:  
$$
\text{Recall@1000} = \frac{N_r}{N_t}
$$  
where $N_r$ is the number of relevant documents retrieved within the top 1000 results, and $N_t$ is the total number of relevant documents.

\subsection{Details of the word evaluation: task description and metrics}\label{sec_appendix:task_description}

\begin{figure*}[!h]
    \centering
    \includegraphics[width=0.9\textwidth]{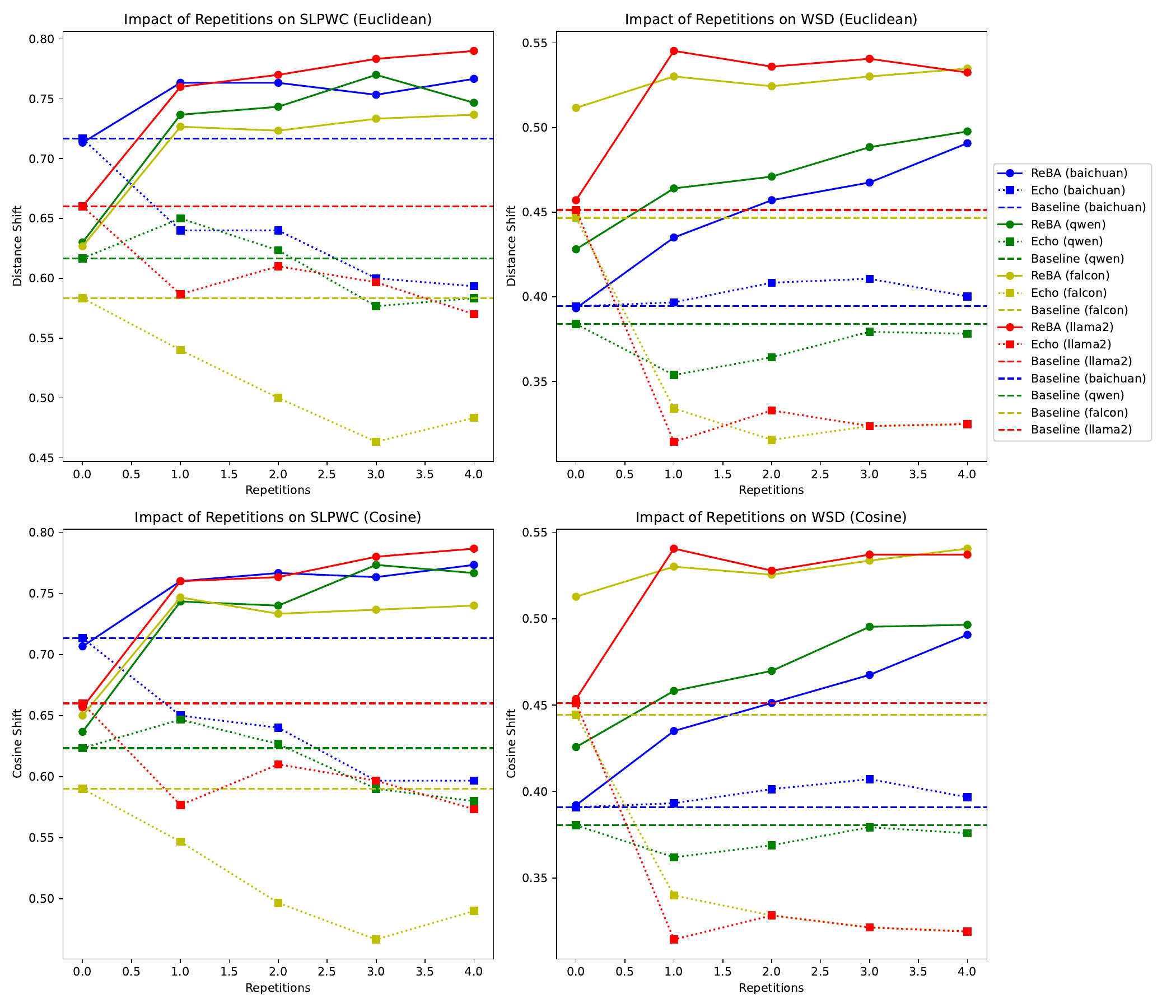}
    \caption{Performance on SLPWC and WSD tasks using Euclidean and Cosine distances to evaluate \emph{word embeddings}, it shows that our results still hold under different distances, }
    \label{fig:word_cos_results}
\end{figure*}

We consider a four-choice question in the word evaluation, where each question has four options. For each question, we extract the word embeddings corresponding to the target word $w_i$ using different methods (ReBA, Echo, Classical) (see Table \ref{tab:embedding_methods}). After obtaining the four word embeddings, we calculate the pairwise Euclidean distances between the four vectors and select the word with the largest sum of distances to the other three vectors as the answer. The Euclidean distance between two vectors $\mathbf{u}$ and $\mathbf{v}$ is:  
$$
d(\mathbf{u}, \mathbf{v})=\|\mathbf{u} - \mathbf{v}\|
$$
In addition, we also consider the cosine distance, which measures the similarity between vectors as: 
The results using cosine distance are shown in figure \ref{fig:word_cos_results}.

\end{document}